\documentclass{article}

\usepackage{microtype}
\usepackage{graphicx}
\usepackage{subfigure}
\usepackage{booktabs} 

\usepackage{hyperref}

\usepackage[accepted]{icml2025}

\usepackage{amsmath}
\usepackage{amssymb}
\usepackage{mathtools}
\usepackage{amsthm}

\usepackage[capitalize,noabbrev]{cleveref}

\usepackage{multirow}
\usepackage{multicol}
\usepackage{adjustbox}

\theoremstyle{plain}

\theoremstyle{definition}

\theoremstyle{remark}

\usepackage[textsize=tiny]{todonotes}
\usepackage{bbding}
\usepackage{xcolor, soul}
\definecolor{light-gray}{gray}{0.85} 
\usepackage{CJKutf8}
\newcommand{\ZH}[1]{\begin{CJK}{UTF8}{gbsn}#1\end{CJK}}

\begin{document}

\twocolumn[
\icmltitle{LUCY: Linguistic Understanding and Control Yielding Early Stage of \emph{Her}}



\icmlsetsymbol{equal}{*}
\icmlsetsymbol{lead}{\(\spadesuit\)}

\begin{icmlauthorlist}
\icmlauthor{Heting Gao}{equal,youtu}
\icmlauthor{Hang Shao}{equal,youtu}
\icmlauthor{Xiong Wang}{equal,youtu}
\icmlauthor{Chaofan Qiu}{youtu}
\icmlauthor{Yunhang Shen}{youtu}
\icmlauthor{Siqi Cai}{youtu,pk}
\icmlauthor{Yuchen Shi}{youtu}
\icmlauthor{Zihan Xu}{youtu}
\icmlauthor{Zuwei Long}{youtu}
\icmlauthor{Yike Zhang}{youtu}
\icmlauthor{Shaoqi Dong}{nj}
\icmlauthor{Chaoyou Fu}{nj}
\icmlauthor{Ke Li}{lead,youtu}
\icmlauthor{Long Ma}{youtu}
\icmlauthor{Xing Sun}{youtu}
\end{icmlauthorlist}

\icmlaffiliation{youtu}{Tencent Youtu Lab, China}
\icmlaffiliation{pk}{Peking University, China}
\icmlaffiliation{nj}{Nanjing University, China}

\icmlcorrespondingauthor{Ke Li}{tristanli@tencent.com}
\icmlcorrespondingauthor{Long Ma}{malonema@tencent.com}
\icmlcorrespondingauthor{Xing Sun}{winfredsun@tencent.com}

\icmlkeywords{Machine Learning, ICML}

\vskip 0.3in
]



\printAffiliationsAndNotice{\icmlEqualContribution\icmlProjectLead} 

\begin{abstract}
The film \emph{Her} features Samantha, a sophisticated AI audio agent who is capable of understanding both linguistic and paralinguistic information in human speech and delivering real-time responses that are natural, informative and sensitive to emotional subtleties. 
Moving one step toward more sophisticated audio agent from recent advancement in end-to-end (E2E) speech systems, we propose LUCY, a E2E speech model that (1) senses and responds to user's emotion, (2) deliver responses in a succinct and natural style, and (3) use external tool to answer real-time inquiries.
Experiment results show that LUCY is better at emotion control than peer models, generating emotional responses based on linguistic emotional instructions and responding to paralinguistic emotional cues.
Lucy is also able to generate responses in a more natural style, as judged by external language models, without sacrificing much performance on general question answering. Finally, LUCY can leverage function calls to answer questions that are out of its knowledge scope. Demos are available at \url{https://github.com/VITA-MLLM/LUCY}.

\end{abstract}

\section{Introduction}
\begin{table*}[ht!]
\caption{Functionality of Recent Speech Models.}
\label{tab:funcs}
\vskip 0.15in
\begin{center}
\begin{footnotesize}
\begin{sc}
\begin{tabular}{lccccccr}
\toprule
\multirow{2}{*}{\textbf{Model}} & \multirow{2}{*}{\textbf{E2E}} & \multicolumn{2}{c}{\textbf{Emotion Control}}   & \multirow{2}{*}{\textbf{Naturalness}} & \multirow{2}{*}{\textbf{Function Calling}} \\\cmidrule(lr){3-4}
                     &  & Linguistic & Acoustic  &  &  \\
\midrule
\textbf{Llama-Omni}  & \XSolidBrush& \XSolidBrush & \XSolidBrush & \Checkmark   & \XSolidBrush \\
\textbf{Freeze-Omni} & \XSolidBrush& \XSolidBrush & \XSolidBrush & \XSolidBrush & \XSolidBrush \\
\textbf{MinMo}       & \XSolidBrush& \Checkmark   & \XSolidBrush & \Checkmark   & \XSolidBrush \\
\textbf{Mini-Omni}   & Parallel    & \XSolidBrush & \XSolidBrush & \XSolidBrush & \XSolidBrush \\
\textbf{Moshi}       & Parallel    & \Checkmark   & \XSolidBrush & \Checkmark   & \XSolidBrush \\
\textbf{GLM-4-Voice} & Interleaved & \Checkmark   & \XSolidBrush & \Checkmark   & \XSolidBrush \\\midrule
\textbf{LUCY}        & Parallel    & \Checkmark   & \Checkmark   & \Checkmark   & \Checkmark   \\
\bottomrule
\end{tabular}
\end{sc}
\end{footnotesize}
\end{center}
\vskip -0.2in
\end{table*}

The film \emph{Her} portrays a near-future scenario where artificially intelligent audio agents can provide users with human-like interactions, delivering real-time responses that are natural, informative, and responsive to user's emotion.
Recent advances in large language model based audio agent systems have moved us one step closer toward the realization of such intelligent audio agents. For example, close-sourced GPT-4o exhibits its ability to answer users' questions in realtime while providing coherent and relevant responses. Recently, many open-sourced E2E systems~\citep{ji2024wavchat} have been proposed, achieving similar levels of performance in terms of latency and coherence.

Despite the architectural advancements of E2E speech systems, we believe that existing systems still lack three key features necessary for a more human-like interaction: (1) emotion control, (2) naturalness, and (3) informativeness.

First, emotion control involves the system's ability to recognize emotions in users' queries and appropriately respond with empathetic and responsive responses. For example, if the user asks the audio agent in an afraid voice, the agent should detect that emotion in speech and respond with a calm and reassuring voice to ease the user's concerns.
Second, naturalness requires the audio agent to generate human-like natural responses, which not only address user needs, but also support and encourage further conversations. Existing large language models tend to generate responses that are verbose and overly formal, often presented in bullet points or incorporating equations and formulas~\citep{fang2024llama,defossez2024moshi,wang2024full}. While precise and detailed in text, such formats are not ideal for spoken interactions, as lengthy and repetitive responses can obscure the main points, making it difficult for users to grasp the essential information. 
Finally, informativeness requires the system to stay updated on the latest knowledge, to provide accurate and contextually relevant responses. To address this, we implement function calling to connect LUCY to external tools, enabling real-time information retrieval and integration from online resources.

We would like to improve E2E speech systems on the aforementioned three aspects. For this purpose, we propose LUCY, an end-to-end speech model trained on data curated to these three aspects. Specifically, we generate text data of $1.0$ million common-purpose conversations (AudioQA-1.0M), $100$k emotional conversations, $60$k natural conversations, and $26$k function-calling conversations, using large language models~\cite{yang2024qwen2} and synthesize the corresponding speech using an in-house TTS system.

LUCY follows the architecture of Mini-Omni~\citep{xie2024mini}, for its fully end-to-end architecture design, which generates speech and text responses in parallel, thus enabling a lower first token delay than interleaved designs. This design also imposes fewer assumptions on speech-text data compared to architectures such as Llama-Omni~\citep{fang2024llama}. 
In order for LUCY to handle multi-round conversations and generate responses with function calls that are silent in speech, we make two modifications to Mini-Omni's batch-parallel decoding method~\citep{xie2024mini}. 
For multi-round conversations, we discard speech responses and save only text responses as contextual history for the generation of the next-round answer. For function calls that are silent in speech, we decode audio and text in separate items with different stopping conditions to prevent the function calls in text responses from being pronounced. Training data are adjusted accordingly to include separate speech and text versions.

In summary, we propose LUCY, an E2E large speech model that enhances emotion control, naturalness, and informativeness by leveraging carefully curated synthetic training data. 
We refine the training and decoding pipelines to handle the added complexities of multi-round interactions and function-call data. Additionally, we develop objective metrics to evaluate the effectiveness of the proposed approach.

\section{Related Works}

Large language models~\cite{achiam2023gpt,touvron2023llama,jiang2023mistral,yang2024qwen2} trained on massive text corpora have demonstrated powerful natural language understanding and generation capability, significantly transforming human-computer interactions. Extending the power to other modalities has since become the next step to explore. Speech, in particular, has emerged as a natural direction to explore due to its central role in daily human communication. A simple baseline is a cascaded ASR-LLM-TTS system with each module separately optimized~\cite{huang2024audiogpt}. Such systems can suffer from high response delays~\cite{ji2024wavchat} because each module requires independent decoding, and reducing latency typically demands substantial effort. Besides, errors from one module can propagate to the subsequent modules in the cascade, further impacting overall performance.

Attempts have been made to integrate the separate modules to avoid the aforementioned disadvantages. Instead of using ASR transcripts as inputs to LLM, several studies~\citep{kong2020panns,chu2023qwen,chu2024qwen2,das2024speechverse} connect the ASR encoder directly to LLM using an trainable encoder adapter. This approach provides richer speech information than pure text. However, a separate TTS is still required to synthesize speech responses from the LLM's text responses. 

To enrich the information fed into TTS systems, one research direction input the languages models' hidden states to the TTS modules. Llama-omni~\citep{fang2024llama} uses a non-autoregressive (NAR) Transformer as TTS module to predict discrete speech tokens from the up-sampled hidden states. This module is optimized  using connectionist temporal classification (CTC) loss. Freeze-omni~\citep{wang2024freeze}
keeps the language model frozen to maximally preserve its language ability and uses a combination of autoregressive and NAR speech decoders that predict speech token from LM hidden states and text embeddings. 
Minmo~\citep{chen2025minmo} combines language models with an autoregressive CosyVoice2 decoder~\cite{du2024cosyvoice} with the former modeling purely text and the latter modeling interleaved speech-text sequences. The hidden state from the language model is input to the CosyVoice2 decoder to predict speech tokens autoregressively, which are subsequently input to a vocoder to synthesize waveforms.

Fully end-to-end systems further integrate the TTS function into LLM; the LLM is responsible for generating not only the text responses but also speech responses, with the latter usually in the form of discrete speech tokens derived from self-supervised speech models~\citep{lakhotia2021generative} or neural audio codecs according to \citet{defossez2022high}. 
These systems can be categorized into two paradigms~\citep{chen2024slam}: interleaved audio-text modeling and parallel audio-text modeling. In the interleaved modeling paradigm~\citep{nguyen2025spirit,zeng2024glm,zeng2024scaling} the text tokens and speech tokens are interleaved into one sequence and the model alternate between predicting text and speech token. In the parallel modeling paradigm, the model predict text and possibly multiple speech sequences simultaneously, using $N$ extra language modeling head for the speech sequences. Moshi~\citep{defossez2024moshi} trains Mimi, an autogressive neural codec with eight codebooks, resulting in $N=8$. Mini-omni uses SNAC codec with seven codebooks, thus $N=7$. Slam-omni~\citep{chen2024slam} split the speech tokens extracted using CosyVoice tokenizer~\citep{du2024cosyvoice} into three groups, resulting in $N=3$. Compared to interleaved modeling, parallel modeling greatly compress the speech tokens into shorter sequences and can model speech tokens of higher bit rates. However, this approach theoretically places a greater load on the LLM, and its impact on the LLM's linguistic performance compared to interleaved approach remains an area for further investigation.

We believe that a good audio agent should excel in three dimensions: emotion control, naturalness, and informativeness (Function-Calling).
Emotion control can be subdivided into two aspects: generating emotional speech explicitly or implicitly instructed by the query content (Linguistic Emotion Control) and sensing the user's emotions conveyed in speech attributes such as volume, pitch, and prosody, to respond appropriately (Acoustic Emotion Control).
A summary of recent speech models is presented in Table~\ref{tab:funcs}, showing that these models typically address one or two of these aspects. LUCY is the first model to achieve improvements across all three dimensions.

\section{Methods}
 
\subsection{Architecture}

\begin{figure}[t!]
\vskip 0.1in
\begin{center}
\centerline{\includegraphics[width=\columnwidth,trim={0.75cm 0.7cm 2cm 0.5cm},clip]{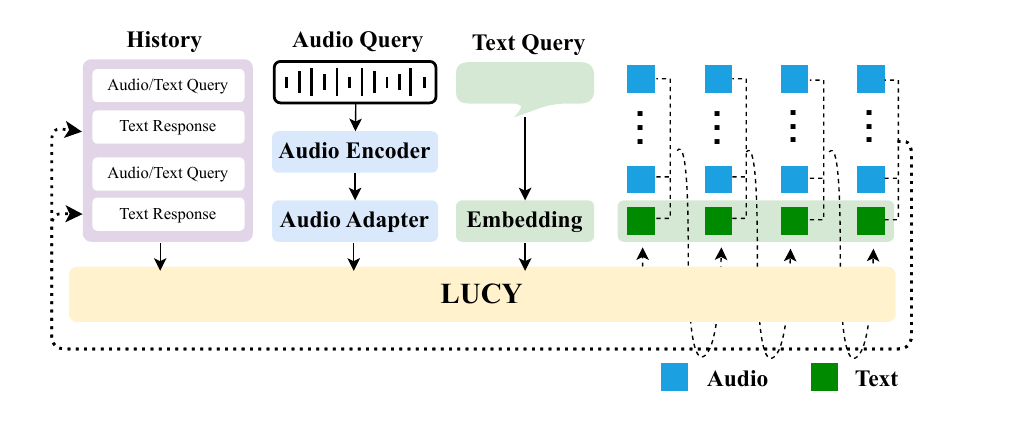}}
\caption{Architecture overview.}
\label{fig:arch}
\end{center}
\vskip -0.4in
\end{figure}

Denote the user's input as $X\in\mathcal{X}$, where $\mathcal{X}$ is the joint space for text and speech embeddings. Denote the model's text response and speech response as $Y^T\in\mathcal{V}$ and $Y^A\in\mathcal{U}$ respectively, where $\mathcal{V}$ is the text vocabulary set and $U$ is the speech codec vocabulary set. Denote the dataset as $\mathcal{D}=\{(X_i,Y_i^T,Y_i^A)\}_{i=1}^N$, where $N$ is the number of samples in the dataset. The negative log-likelihood $\mathcal{L}$ over the dataset $\mathcal D$ can be modeled as 

\begin{equation}
    \mathcal{L}=\sum_{i=1}^N\sum_{t=1}^{l_i}\log P(Y^T_{i,t},Y^A_{i,t}|Y^T_{i,<t},Y^A_{i,<t},X_i),
\end{equation}

where $l_i$ is the maximum number of tokens of the output text $Y^T_i$ and the output speech $Y^A_i$ of the $i$-th sample.

The overview of LUCY's architecture is shown in Figure~\ref{fig:arch}, which consists of an audio encoder, an audio adapter, a language model backbone and eight language-modeling heads. Detailed descriptions of each component is provided as follow.


\textbf{Backbone Model}\quad 
The backbone of LUCY is a decoder-only Transformer based language model for which we use Qwen2-7B-Instruct~\citep{chu2024qwen2}. This backbone model processes users' queries, whether in speech or text, and generates responses in both modalities.

\textbf{Audio Encoder and Adapter}\quad 
We use a Transformer-based audio encoder~\citep{wang2024freeze} to encode input speech into continuous features, which has $350$M parameters consisting of multiple downsampling convolutional layers ($4$x downsampling) and $24$ Transformer blocks with a hidden size of $1024$. We extract $80$-bin Mel-filter bank features from $16$ kHz raw waveform using a $25$-ms shifting window with a hop size of $10$ms. The audio encoder then encodes the Mel features into speech features with a frame rate of $25$Hz.
We use a convolutional layer based audio adapter to align the audio modality with the backbone language model, which further downsamples the speech features into $12.5$ Hz.

\textbf{Codec}\quad 
Similar to Mini-omni~\citep{xie2024mini}, we use SNAC codec~\citep{siuzdak2024snac} to encode the output speech into discrete tokens with $7$ codebooks and a total token rate of $82$ Hz. 
We follow the parallel modeling paradigm~\citep{copet2024simple} to decode text and speech tokens simultaneously. 
Combining speech and text tokens results in a total of $8$ layers of targets, so we have eight language-model heads to predict one text token and seven audio token at each decoding step. One token delay is applied to each layer of audio tokens for better generation quality, as shown in Figure~\ref{fig:arch}.

\textbf{Emotion Control}\quad
\begin{figure}[t!]
\vskip 0.1in
\begin{center}
\centerline{\includegraphics[width=\columnwidth,trim={0.75cm 0.5cm 0.5cm 0.cm},clip]{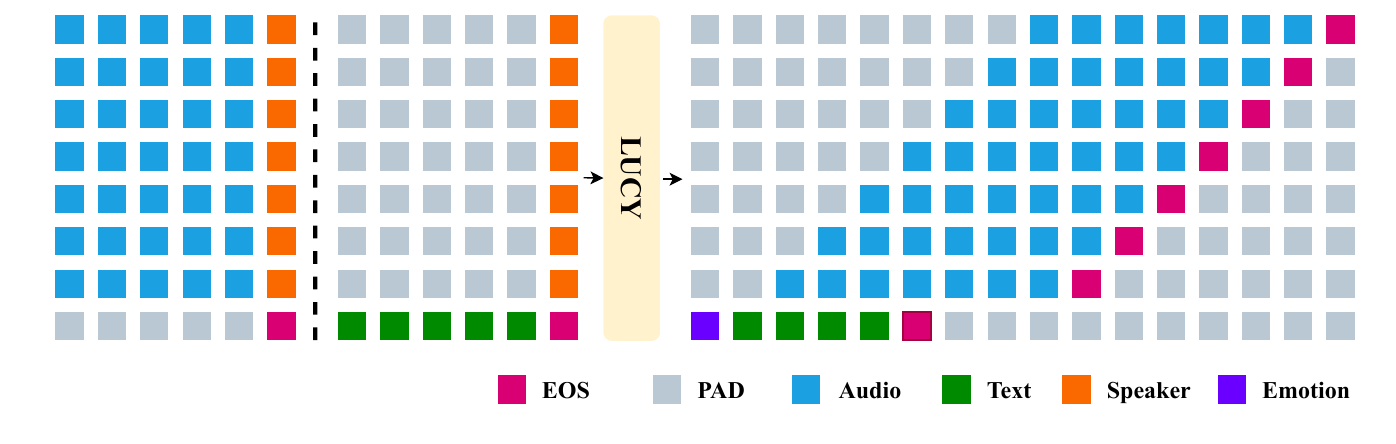}}
\caption{Illustration of Emotion and Speaker Tokens.}
\label{fig:emo}
\end{center}
\vskip -0.4in
\end{figure}
We use special tokens to control the speaker and the emotion of the generated response.
LUCY is trained on two distinct voices, one male and one female. We append a speaker token after user inputs $X$ to select the desired speaker. To control the emotion in the response, we inset additional emotion tokens before the text, so LUCY learns to first predict the emotion token and then responses. The former thus serves as emotion context which guide the generation of the subsequent audio tokens. The emotion and speaker tokens are inserted as shown in Figure~\ref{fig:emo}.

\textbf{Multi-Round Conversation History}\quad 
We train LUCY using multi-round conversation data. In this setup, queries and responses from previous rounds are appended to the conversation history, which then serves as context for generating responses in subsequent rounds. We incorporate query inputs into the history irrespective of their modality. However, to minimize the length of the history while avoiding information loss, we only append the text version of responses, excluding the generated speech responses. The training data are adjusted accordingly, where all historical responses in the training data for multi-round conversations are maintained exclusively in text form.

\textbf{Function Calls}\quad 
\begin{figure}[t!]
\begin{center}
\centerline{\includegraphics[width=\columnwidth,trim={0.75cm 0.5cm 1cm 0.cm},clip]{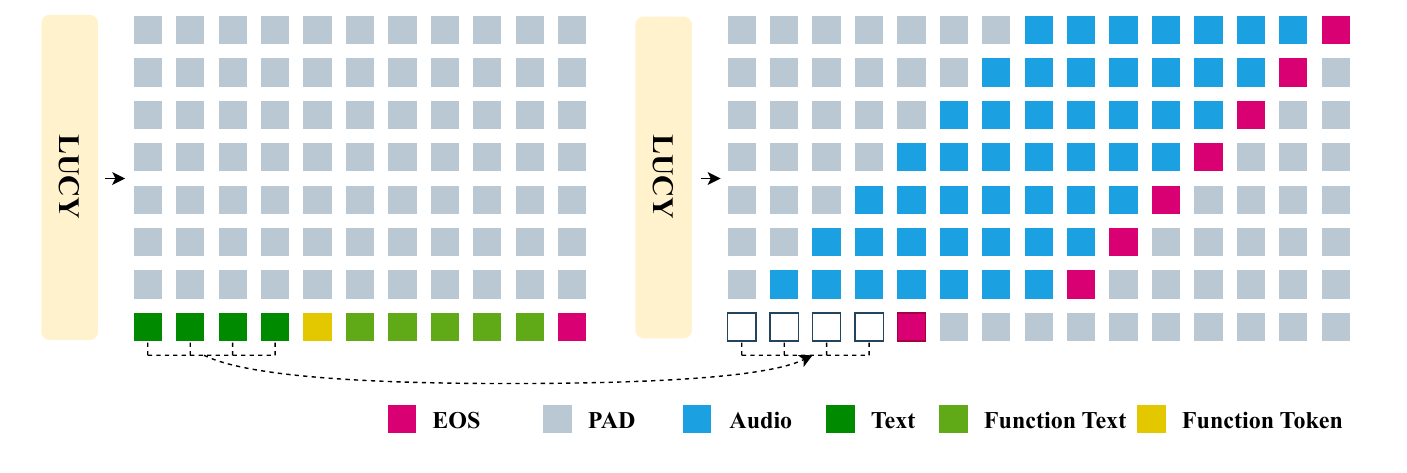}}
\caption{Batch Parallel Decoding for Function-Call Samples.}
\label{fig:func}
\end{center}
\vskip -0.3in
\end{figure}
Conversations involving function calls contain responses that are partially silent. For example, when the user queries about the current weather, the assistant should generate a voiced query-acknowledging response and an unvoiced function call to the weather API. 
To handle function calling conversations, we make two variants of one function-call sample: one text version where the model generates both voiced and unvoice text response and one speech-text version where the model generates voiced text response and its corresponding speech response. During inference We adapt 
batch-parallel decoding to generate function-calling responses. Batch-parallel decoding expands a single input to a batch size of two: the first sample generates text-only response and the second sample generates both text and speech response. The generated text tokens from the first sample are of better quality than those from the second and are used to substitute the latter in the second sample.
In the case of function-call response, we use the first text-only sample to generate the voiced acknowledging response and the unvoice function-calling response and use the second text-speech sample to generate the voiced acknowledging speech with its text tokens replaced by those from the first sample. This process is illustrated in Figure~\ref{fig:func}.

\textbf{Training Pipeline}\quad
The training of LUCY consists of three stages. In the first stage, we freeze the language model and train the audio encoder and adapter from scratch on a large in-house ASR dataset. This dataset comprises approximately $110$k hours of speech data, split between about $60$k hours of Chinese and $50$k hours of English. The primary goal during this stage is for the encoder and adapter to effectively transform speech waveforms into features that are comprehensible to the language model.

In the second stage, we freeze the audio encoder and adapter and train the language model on AudioQA-1.0M, a large spoken dialog dataset consisting of one million conversations generated using Qwen2.5-72B~\cite{yang2024qwen2} and a flow-matching based TTS system, so the model learn to generate speech and text response following the parallel text-speech modeling paradigm.

The third stage involves further fine-tuning the language model on the synthesized conversation dataset, with a specific focus on enhancing three key aspects: emotion control, function-calling, and naturalness. These aspects are elaborated upon in subsequent sections.

\subsection{Text Generation}

\subsubsection{Emotional Understanding and Control}
Due to the fact that current mainstream E2E large models for voice interaction~\cite{xie2024mini,chen2024slam} do not enrich the emotional quality of voice output, the agent's responses often appear monotonous and dull. To make the voice interaction experience more realistic and engaging, we propose two methods for emotional enhancement to make LUCY's voice interaction emotions more lifelike and human-like.

\textbf{Linguistic Emotion Control}\quad
\label{sec:emoctrl}
To enrich the emotional output of speech, we refer to mainstream emotion recognition models~\cite{chen2022wavlm,ma2023emotion2vec}, open-source datasets~\cite{poria2018meld,zhao2022m3ed}, and benchmarks~\cite{ma2024emobox} to select seven types of emotions, which are \emph{Joy}, \emph{Sadness}, \emph{Fear}, \emph{Anger}, \emph{Surprise}, \emph{Disgust}, and \emph{Neutral}. Additionally, to provide better emotional responses to certain emotions, we introduce \emph{Sorry} emotion, resulting in an eight-emotion label system.

Linguistic emotion control can be divided into explicit and implicit control. Explicit control is where the user explicitly specifies the emotion of the reply, for example, ``Explain the football match in a cheerful voice." The data for generating explicit instruction conversations comes from the common-purpose dialogue dataset, AudioQA-1.0M, where We filter out conversations with the input and output lengths than $1,000$ tokens and then randomly sample $40$k from the remaining. For each round of responses, we classify the emotion tags of the voice replies and regenerated the input instructions that controls the response emotion using Qwen2.5-72B model. Doing so ensures that the instructions explicitly reflect the emotion of replies. After format checking, we obtain around $36$k multi-turn dialogue datasets in both Chinese and English with explicit emotion control.

Implicit control means that the input instruction does not specify the emotion of the reply, and the reply emotion must be inferred from the semantic emotion in the instruction. 
For example, if a user says ``I'm sick today'', then the agent may reply, ``That sounds awful, get well soon!" with a sorry emotion. The implicit control data are sampled from the AudioQA-1.0M dataset, amounting to 45K samples. Based on the semantic emotion of the query, we generate a reasonable voice emotion response tag and rewrite the response so its content better match the generated emotion tag. After filtering out mal-format conversations, we obtain over $43$K multi-turn dialogue datasets in both Chinese and English with implicit emotion generation control.

\textbf{Acoustic Emotion Control}\quad
Both explicit and implicit control generate emotional responses based on the semantic emotional information from the query, without considering the acoustic emotional information in the user's speech. To enable LUCY to take into account not only the semantic information but also acoustic paralinguistic cues when 
emotional responses, we construct a multi-turn acoustic-emotion dialogue dataset. This dataset is created by sampling $65$K multi-turn dialogues from AudioQA-1.0M utilizing Algorithm~\ref{alg:EARC}.
Specifically, for each dialogue turn's query, an emotion is randomly selected from the emotional label system as the query's vocal emotion, while the semantic emotion of the query is obtained using an LLM (Qwen2.5-72B). If the acoustic and semantic emotions are the same, the query's composite emotion is either one of them; if they differ, we check for neutrality. If neutrality is present, the composite emotion is the non-neutral one; otherwise, the acoustic and semantic emotions may be different or even contradictory. Therefore, the composite emotion is a weighted average of the acoustic and semantic emotions. With $\alpha$ as the weighting coefficient, we set $\alpha=0.7$  to make vocal emotion perception more prominent.

\begin{footnotesize}
\begin{algorithm}[tb]
   \caption{Emotion-Aware Response Constructor}
   \label{alg:EARC}
\begin{algorithmic}
   \STATE {\bfseries Input:} $emoList$, List of all $8$ emotion labels.\\
   \STATE {\bfseries Input:} $C$, Natural Multi-turn Conversations Without Vocal Emotion Labels.\\
   \STATE {\bfseries Output:} $C'$, Multi-turn Emotion-aware Conversations. \\
   \STATE Initialize $history$ = $Empty$
   \FOR{$query$ {\bfseries in} $C$}
    \STATE $acoustic\_emo$ = {\bfseries Random}($emoList$)
    \STATE $semantic\_emo$ = LLM$(query, emoList)$
   \IF{$acoustic\_emo == semantic\_emo$}
    \STATE $query\_emo$ = $acoustic\_emo$
    \ELSE
      \IF{$acoustic\_emo$ $\neq$ $Neutral$ $\AND$ $semantic\_emo$ $\neq$ $Neutral$}
        \STATE $query\_emo$ = $\alpha$ $*$ $acoustic\_emo$ + (1 - $\alpha$) $*$ $semantic\_emo$
      \ELSE
         \STATE $query\_emo$ = ($\lnot$ $Neutral$)\\($acoustic\_emo, semantic\_emo$)
      \ENDIF
      \STATE $answer\_emo$, $answer$ = LLM$(query\_emo,$ \\ $query, history)$ \\
      \STATE $history$ += ($query$, $answer$)
   \ENDIF
   \STATE $C'$ add $(query\_emo, query, answer\_emo, answer)$
   \ENDFOR
   
\end{algorithmic}
\end{algorithm}
\end{footnotesize}

\subsubsection{Function call}

We generate function calls by adapting existing function calling datasets such as WebPCM~\citep{qin2023webcpm} and DevOpsEval~\citep{DevOpsEval}. Specifically, we filter out functions that are not suitable for oral interactions. About $40$ functions calls remains, including general Google search, weather search, news search, etc. We then use Qwen2.5-72B model to rewrite the agent responses to make them less formal and better suited for oral interaction. Finally, we implement a phased response mechanism to address latency issues associated with parameter generation for tool invocation, where the model provides concise transitional messages prior to completing parameter generation, such as ``Sure, I can look that up for you'' to acknowledge the receipt the query. Given that the speed of parameter generation and tool invocation is significantly faster than that of speech synthesis, this mechanism effectively bridges the perceived waiting time, thereby enhancing interaction fluidity and improving user satisfaction. The true response addressing user's inquiry follows the acknowledgment response after the function-calling results are returned.

\begin{table*}[ht!]
\caption{Emotion Control Results. LEC stands for Linguistic Emotion Control and AEC stands for Acoustic Emotion Control.}
\label{tab:SE}

\begin{center}
\begin{footnotesize}
\begin{sc}
\begin{adjustbox}{max width=\textwidth}
\begin{tabular}{lcccccccccr}
\toprule
\textbf{Model} & \textbf{Mode} & \textbf{Lang} & \textbf{Neutral} & \textbf{Joy} & \textbf{Anger} & \textbf{Fear} & \textbf{Disgust} & \textbf{Sadness} & \textbf{Surprise} & \textbf{Sorry}  \\
\midrule
\multirow{4}{*}{GLM-4-Voice} & LEC & ZH & 97.13 & 50.17 & 11.13 & 1.10 & 0 & 27.20 & 62.68 & 0  \\
& LEC & EN & 64.05 & 78.23 & 11.19 & 5.79 & 5.81 & 57.48 & 20.51 & 0.57 \\
& AEC & ZH & 93.76 & 15.75 & 0 & 0 & 0 & 1.29 & 3.87 & 0 \\
& AEC & EN & 89.42 & 21.67 & 0 & 0 & 0 & 5.88 & 0 & 0.93 \\
\midrule
\multirow{4}{*}{LUCY} & LEC & ZH & 99.45 & 96.18 & 92.94 & 88.12 & 73.54 & 85.47 & 84.32 & 88.69 \\
& LEC & EN & 97.31 & 95.63 & 90.23 & 86.54 & 68.79 & 81.76 & 79.18 & 84.27 \\
& AEC & ZH & 79.54 & 66.67 & 55.84 & 49.52 & 28.57 & 49.64 & 43.04 & 40.54  \\
& AEC & EN & 80.27 & 60.46 & 46.67 & 30.77 & 12.50 & 41.10 & 24.44 & 33.01 \\
\bottomrule
\end{tabular}
\end{adjustbox}
\end{sc}
\end{footnotesize}
\end{center}
\vskip -0.2in
\end{table*}

\subsubsection{Natural Conversation}
The development of natural and human-like conversational AI requires training data that reflects genuine human interaction patterns. While existing LLMs can function as capable assistants through text-based interactions, their responses often lack the natural flow essential for spoken dialogue, particularly when handling technical elements such as code snippets, URLs, or mathematical formulae. To address this limitation, we implement a comprehensive three-phase data processing pipeline.
First, we filter the initial dataset to remove entries containing extensive technical elements and responses exceeding 1,000 tokens. Second, we employ a model-based rewriting process to improve the colloquial nature of the filtered responses while preserving their semantic meaning. Finally, we implement a quality control phase where each conversation is evaluated through a multi-dimensional scoring system, assessing factors such as topic coherence, tone appropriateness, and overall fluency on a scale of 1-5.
Through this systematic approach, we refine our original dataset of $600$k conversation pairs down to approximately $60$k high-quality, colloquial entries; only conversations scoring $4$ or above across all evaluation dimensions are retained in the final dataset.

\subsection{Audio Generation}
We generates the spoken conversation data using an internal emotion control zero-shot TTS system. The specific speech generation procedure for the user questions and agent answers are detailed as  follows.

\textbf{Generation for questions from user}\quad
In the training process, user's question speech should have a sufficient range of speaker timbres and styles so that the model possesses strong acoustic robustness. Against this backdrop, we use the audio from the ASR corpus as the reference speech for the zero-shot TTS system, ensuring that the generated data also have sufficient diversity. For questions with emotions, we generate weak emotion labels for the data in the ASR corpus, select the entries with higher confidence scores, and combine the emotion control ability in the zero-shot TTS system to obtain the question speech with emotions.

\textbf{Generation for answers from agent}\quad
Similar to the generation of question speech, the answer speech is also generated by zero-shot TTS system. The difference is that the reference speech used here is sourced from the high-quality, hypernatural, multi-emotional TTS data of limited speakers, to ensure that the agent can output stable and high-quality speech. We select two speakers as the reference speaker for LUCY, one male and one female.

\begin{table*}[t]
\caption{Function Call Results.}
\label{tab:FC}
\begin{center}
\begin{small}
\begin{sc}
\begin{adjustbox}{max width=\textwidth}
\begin{tabular}{lcccccccr}
\toprule
\multirow{2}{*}{\textbf{Model}} & \multirow{2}{*}{\textbf{Modality}} & \multicolumn{4}{c}{\textbf{Tool Selection}} & \textbf{Parameter Extraction} & \textbf{Response Generation} & \textbf{Overall} \\
& & ACC & F1 & Precision & Recall & ACC & ACC & ACC \\
\midrule
Qwen2-7B & Text & 57.50 & 69.75 & 54.14 & 98.00 & 97.00 & 72.00 & 35.00 \\
\midrule
\multirow{2}{*}{LUCY} & Audio & 94.00 & 94.06 & 95.00 & 93.14 & 65.00 & 46.00 & 63.00 \\
& Text & 95.05 & 96.00 & 94.12 & 95.00 & 94.00 & 58.00 & 74.50 \\
\bottomrule
\end{tabular}
\end{adjustbox}
\end{sc}
\end{small}
\end{center}
\vskip -0.1in
\end{table*}

\begin{table}[h]
\caption{Case Study of Responses From Qwen2-7B and LUCY.}
\label{tab:casestudy}
\vskip 0.15in
\begin{center}
\begin{footnotesize}
\begin{tabular}{l|cccc}
\toprule
\textbf{Question} & \multicolumn{4}{l}{\begin{footnotesize}\ZH{跑步时除了穿好鞋，还需要注意些什么？}\end{footnotesize}} \\\midrule
\multicolumn{5}{p{7.7cm}}{  
    \textbf{Qwen2-7B}: \begin{footnotesize}\ZH{跑步时还需要注意以下几点：
1. 穿着合适的运动服装，以保持舒适和散热。
2. 确保鞋子的鞋底足够坚固，以提供足够的支撑。
3. 确保鞋子的鞋带松紧适度，以保持脚部的舒适。
4. 确保鞋子的鞋底有足够的抓地力，以提供足够的抓地力。
5. 确保鞋子的鞋底有足够的厚度，以提供足够的缓冲。
6. 确保鞋子的鞋底有足够的弹性，以提供足够的弹性。
7. 确保鞋子的鞋底有足够的耐磨性，以提供足够的耐磨性。}\end{footnotesize}
}  \\\midrule
\multicolumn{5}{p{7.7cm}}{
    \textbf{LUCY}: \begin{footnotesize}\ZH{跑步时需要注意呼吸、保持节奏、保持身体平衡，还要注意饮食和休息哦。}\end{footnotesize}
}  \\\midrule\midrule

\textbf{Question}&\multicolumn{4}{l}{\begin{footnotesize}\ZH{退款流程是怎样的？}\end{footnotesize}} \\\midrule
\multicolumn{5}{p{7.7cm}}{
    \textbf{Qwen2-7B}: \begin{footnotesize}\ZH{
    退款流程如下：
1. 首先，您需要联系我们的客服，说明您的退款需求。
2. 客服会根据您的情况，提供退款的详细流程和所需材料。
3. 您需要提供相关的证明材料，如课程购买凭证、课程进度截图等。
4. 客服会根据您的情况，提供退款的详细流程和所需材料。
5. 在确认退款后，您需要按照客服的指示，将退款金额转入指定的账户。
6. 退款完成后，我们会及时通知您退款成功。
以上就是我们的退款流程，希望对您有所帮助。}\end{footnotesize}
}  \\\midrule
\multicolumn{5}{p{7.7cm}}{ 
    \textbf{LUCY}: \begin{footnotesize}\ZH{退款流程是：您申请退款后，我们的课程顾问会与您联系，我们会根据您参加的课程时间，判断是否可以退款，如果是的话，我们会为您办理退款手续。}\end{footnotesize} 
} \\
\bottomrule
\end{tabular}
\end{footnotesize}
\end{center}
\vskip -0.2in
\end{table}

\section{Experiment}

\subsection{Evaluation on Speech Emotion}

\textbf{Linguistic Emotion Control}\quad
In Linguistic Emotion Control~(LEC) mode primarily evaluates a model's ability to generate appropriate emotional speech responses based on the semantic information of the query. As described in Section ~\ref{sec:emoctrl}, it includes both explicit and implicit emotion generation control. For this purpose, we construct a test set of $1500$ emotion generation control entries, utilize emotion2vec~\citep{ma2023emotion2vec} to annotate the emotions of the generated speech, and then compare them with ground truth emotions. Since emotion2vec does not have a \emph{Sorry} emotion candidate, we categorized \emph{Other} as \emph{Sorry}. As shown in Table~\ref{tab:SE}, LUCY achieves relatively stable emotional speech output across various emotions. In contrast, GLM-4-Voice shows deficiencies in expressing emotions like \emph{Disgust}, \emph{Fear}, and \emph{Sorry}. These results demonstrates LUCY model's rich capability in speech emotion control.

\textbf{Acoustic Emotion Control}\quad
In Acoustic Emotion Control~(AEC) mode, we evaluate a model’s ability to generate nuanced emotional speech responses based on both the acoustic and semantic emotions given the user's query. In certain scenarios, such as a user speaking neutrally but with an angry tone, it is also necessary to provide a comforting response. For this purpose, we use Algorithm~\ref{alg:EARC} to construct an acoustic-emotion-control test set where we also use emotion2vec to annotate the emotions of the model's output speech for comparison with the groundtruth emotions. As shown in Table~\ref{tab:SE}, LUCY achieves about $80\%$ accuracy for neutral emotions, which is lower than GLM-4-Voice. The primary reason is that GLM-4-Voice, in LER mode, cannot perceive the acoustic emotions of the user's query, resulting in most responses being neutral. On the other hand, LUCY can trigger, with a certain probability, response for various other emotions, though it performs relatively weak on \emph{Disgust}, which may be attribted to scarcity of \emph{Disgust} emotion in both training data and in real-world scenarios.

\subsection{Evaluation on Function Call}
Function-call design enables audio agents to utilize external knowledge bases for real-time queries, with a major focus on integrating the capabilities of the Google search engine to provide users with the latest information. We have constructed a test set of $200$ dialog scenarios suitable for spoken interactions, with a $1:1$ ratio of positive and negative examples, to evaluate LUCY's ability in tool selection, parameter extraction, and response generation. Parameter extraction and response generation are compared based on Qwen2.5-72B and the standard answers. Parameter extraction mainly assesses the model's ability to identify the keywords in the user's query for tool invocation, while response generation evaluates whether the model can output the correct answers based on the returned results from tool calls. The overall accuracy assesses models' ability to perform three sub-functions all correctly. As shown in Table~\ref{tab:FC}, the input from the speech modality is generally lower in various metrics compared to the text modality, likely due to errors in speech recognition. Compared to Qwen2-7B on Overall accuracy, our model shows significant advantage in handling tool calls even with speech input.

\subsection{Evaluation on Natural Conversation}

We evaluate the naturalness of the generated response on our in-house dataset, AudioChat, of mainly daily-conversations in Chinese. Qwen2.5-72B~\citep{yang2024qwen2} is used to compare the naturalness of the text responses generated by the speech model against those generated by the Qwen2-7B model and to decide which one is more natural. The win rate is presented in Table~\ref{tab:audiochat}. We also experiment with GLM-4-Voice and provide its win rate for comparison. In the row of `S$\rightarrow$T', the text responses used are directly generated by the model while in the row of `S$\rightarrow$S', the text responses are obtained by transcribing the model's speech responses using an external ASR model, for which we use `whisper-medium'~\citep{radford2023robust}. From Table~\ref{tab:audiochat}, we observe that LUCY is able to outperform GLM-4-Voice in terms of naturalness, despite having few parameters. To better understand the naturalness score, we present a case study comparing the responses generated by LUCY models and by Qwen2-7B in Table~\ref{tab:casestudy}. We see that LUCY can generate concise answers while maintaining core information.
\begin{table}[t]
\caption{Natural Conversation Win-Rate Results.}
\label{tab:audiochat}
\vskip 0.15in
\begin{center}
\begin{footnotesize}
\begin{sc}
\begin{tabular}{lcccr}
\toprule
\textbf{Model} & \textbf{Lang} & \textbf{Modality} & \textbf{AudioChat} \\
\midrule
GLM-4-Voice &  ZH & S $ \rightarrow $ T & 53.94  \\
GLM-4-Voice &  ZH & S $ \rightarrow $ S & 47.37  \\
\midrule
LUCY & ZH & S $ \rightarrow $ T & 56.02 \\
LUCY & ZH & S $ \rightarrow $ S & 54.33 \\
\bottomrule
\end{tabular}
\end{sc}
\end{footnotesize}
\end{center}
\vskip -0.1in
\end{table}

\subsection{Evaluation on Spoken Question Answering}
\begin{table*}[t]
\caption{Spoken Question Answering Results.}
\label{tab:SQA}
\vskip 0.15in
\begin{center}
\begin{footnotesize}
\begin{sc}
\begin{tabular}{lcccccr}
\toprule
\textbf{Model} & \textbf{Lang} & \textbf{Modality} & \textbf{\#Params}  & \textbf{Web Questions} & \textbf{Llama Questions} & \textbf{TriviaQA} \\
\midrule
Moshi & EN & S $ \rightarrow $ T & 7B & 26.60 & 62.30 & 22.80 \\
GLM-4-Voice & EN & S $ \rightarrow $ T & 9B & 32.20 &  64.70 &  39.10 \\
LUCY (S2) & EN & S $ \rightarrow $ T & 7B & 29.28 & 57.33 & 27.03 \\
LUCY (S3) & EN & S $ \rightarrow $ T & 7B & 26.67 & 59.67 & 23.26 \\
\midrule

Moshi & EN & S $ \rightarrow $ S & 7B & 9.20 & 21.00 & 7.30 \\
GLM-4-Voice & EN & S $ \rightarrow $ S & 9B & 15.90 & 50.70 & 26.50 \\
LUCY (S2) & EN & S $ \rightarrow $ S & 7B & 25.64 & 54.00 & 22.87 \\
LUCY (S3) & EN & S $ \rightarrow $ S & 7B & 18.22 & 51.00 & 18.22 \\
\bottomrule
\end{tabular}
\end{sc}
\end{footnotesize}
\end{center}
\vskip -0.2in
\end{table*}

\begin{table*}[t]
\caption{Automatic Speech Recognition Results.}
\label{tab:ASR}
\vskip 0.15in
\begin{center}
\begin{footnotesize}
\begin{sc}
\begin{tabular}{lcccccccr}
\toprule
\multirow{2}{*}{\textbf{Model}} & \multicolumn{2}{c}{\textbf{WenetSpeech}} & \textbf{AIshell} & \multicolumn{4}{c}{\textbf{LibriSpeech}} \\\cmidrule(lr){2-3}\cmidrule(lr){4-4}\cmidrule(lr){5-8}
& test\_net & test\_meeting & test & dev-clean & dev-other & test-clean & test-other \\
\midrule
Qwen2-Audio & 7.72 & 8.40 & 1.53 & 1.30 & 3.40 & 1.60 & 3.60 \\
GLM-4-Voice & - & - & 3.02 & -& - & 2.10 & 4.90 \\
VITA & 12.15 & 16.53 & - & 7.57 & 16.57 & 8.14 & 18.41  \\
Freeze-Omni & 11.8 & 13.46 & 2.48 & 4.03 & 9.45 & 3.82 & 9.79 \\
\midrule
LUCY & 8.78 & 10.42 & 2.40 & 3.48 & 7.95 & 3.36 & 8.05 \\
\bottomrule
\end{tabular}
\end{sc}
\end{footnotesize}
\end{center}
\vskip -0.2in
\end{table*}

We evaluate LUCY's question-answering abilities on three publicly available English datasets: Web-Questions~\citep{berant2013semantic}, Llama Quesitons~\citep{nachmani2023spoken} and TriviaQA~\citep{joshi2017triviaqa}. We check if the phrases of groundtruth answers presents in the generated responses and compute the overall presence rate in Table~\ref{tab:SQA}. Similar to Table~\ref{tab:audiochat}, the row `S$\rightarrow$T' uses the text responses directly generated by the model and the row `S$\rightarrow$S' uses transcripts from speech responses using whisper-medium. We provide two versions of LUCY one after stage 2 training (S2) and the other after stage 3 training (S3). We observe that performance of both versions of LUCY in S$\rightarrow$T modality is better than Moshi but is worse than GLM-4-Voice. While stage-3 training enhances LUCY's emotion controling, naturalness, and function calling in Chinese conversations, it degrades LUCY's question answering ability in English to some degree. In S$\rightarrow$T modality, S2 LUCY's performance is better than that of GLM-4-Voice because of a smaller degredation from S$\rightarrow$T to S$\rightarrow$S, which may indicate the speech generated by LUCY is of higher quality.

\subsection{Evaluation on Other Abilities}
\textbf{ASR}\quad
We evaluate LUCY's ASR performance after Stage $1$ training on two Chinese datasets (WenetSpeech and AIshell), and one English dataset (LibriSpeech), and report the results in Table~\ref{tab:ASR}. We follow the original data split of each dataset. The results of Qwen2-Audio, GLM-4-Voice, VITA and Freeze-Omni are partially copied from the their respective works for comparison. We see that the ASR performance of LUCY is comparable to that of VITA and Freeze-Omni which also use a text-based language model as backbone.

\textbf{Full-duplex Accuracy}\quad
\begin{table}[t]
\caption{Classification Performance of Negative Sample Rejection.}
\label{tab:negative}

\vskip 0.15in
\begin{center}
\begin{footnotesize}
\begin{sc}
\begin{adjustbox}{max width=\textwidth}
\begin{tabular}{lccccccccr}
\toprule
& \textbf{Precision} & \textbf{Recall} & \textbf{F1 Score} \\
\midrule
VITA & 0.59 & 0.94 & 0.73 \\
LUCY & 0.98 & 0.98 & 0.98 \\
\bottomrule
\end{tabular}
\end{adjustbox}
\end{sc}
\end{footnotesize}
\end{center}
\vskip -0.25in
\end{table}
In order for the model to have full duplex capability, we additionally synthesize negative speech data that the model should not respond to. The negative data consists of $44$k declarative sentences that does not require a response and $40$k pieces of recordings of environmental sounds. 
We measure the rejection accuracy on a hold-out split of dataset of $400$ samples with positive and negative samples evenly split. The results are shown in Table~\ref{tab:negative}. The F1 score of classifying negative and positive queries is $98\%$, which is higher than that of VITA.

\textbf{Latency}\quad
\begin{table}[t]
\caption{Delay in second (s) of Deployed LUCY.}
\label{tab:latency}
\vskip 0.15in
\begin{center}
\begin{footnotesize}
\begin{sc}
\begin{adjustbox}{max width=\textwidth}
\begin{tabular}{lccccccccr}
\toprule
& \textbf{Total} & \textbf{First Chunk} & \textbf{Per-Step} \\
\midrule
LUCY & 0.4652 & 0.3590 & 0.0227 \\
\bottomrule
\end{tabular}
\end{adjustbox}
\end{sc}
\end{footnotesize}
\end{center}
\vskip -0.25in
\end{table}
We deploy LUCY to a web sever and measure the end-to-end in half-duplex mode where the users manually control the timing of inputting speech and no voice activity detection (VAD) module is involved. The results are presented in Table~\ref{tab:latency}. The reported numbers are averaged over ten samples. The end-to-end latency of the entire system ({\sc Total}) in half-duplex mode is $0.4652$ s under a good network connection. The time cost to generate the first audio chunk ({\sc First Chunk}) is $0.3590$ s. Subtracting the first-chunk cost from total latency, we get an estimation of network transmission cost of $0.1062$ s. The time cost to decode one token is $0.0227$ s. All the time costs are measures on a 296-TFLOPS GPU.

\section{Conclusion}
In this work, we propose LUCY, an E2E large speech model that enhances emotion control, naturalness, and informativeness by leveraging carefully curated synthetic training data. 
We refine the training and decoding pipelines to handle the added complexities of multi-round interactions and function-call data. Additionally, we develop objective metrics and conduct extensive experiment to validate the effectiveness of the proposed approach.

\nocite{langley00}

\bibliography{reference}

\begin{thebibliography}{36}
\providecommand{\natexlab}[1]{#1}
\providecommand{\url}[1]{\texttt{#1}}
\expandafter\ifx\csname urlstyle\endcsname\relax
  \providecommand{\doi}[1]{doi: #1}\else
  \providecommand{\doi}{doi: \begingroup \urlstyle{rm}\Url}\fi

\bibitem[Dev()]{DevOpsEval}
A devops domain knowledge evaluation benchmarks for large language models.
\newblock \url{{https://github.com/codefuse-ai/codefuse-devops-eval}}.
\newblock Accessed: 2025-01-26.

\bibitem[Achiam et~al.(2023)Achiam, Adler, Agarwal, Ahmad, Akkaya, Aleman, Almeida, Altenschmidt, Altman, Anadkat, et~al.]{achiam2023gpt}
Achiam, J., Adler, S., Agarwal, S., Ahmad, L., Akkaya, I., Aleman, F.~L., Almeida, D., Altenschmidt, J., Altman, S., Anadkat, S., et~al.
\newblock Gpt-4 technical report.
\newblock \emph{arXiv preprint arXiv:2303.08774}, 2023.

\bibitem[Berant et~al.(2013)Berant, Chou, Frostig, and Liang]{berant2013semantic}
Berant, J., Chou, A., Frostig, R., and Liang, P.
\newblock Semantic parsing on freebase from question-answer pairs.
\newblock In \emph{Proceedings of the 2013 conference on empirical methods in natural language processing}, pp.\  1533--1544, 2013.

\bibitem[Chen et~al.(2025)Chen, Chen, Chen, Chen, Chen, Deng, Du, Gao, Gao, Gao, et~al.]{chen2025minmo}
Chen, Q., Chen, Y., Chen, Y., Chen, M., Chen, Y., Deng, C., Du, Z., Gao, R., Gao, C., Gao, Z., et~al.
\newblock Minmo: A multimodal large language model for seamless voice interaction.
\newblock \emph{arXiv preprint arXiv:2501.06282}, 2025.

\bibitem[Chen et~al.(2022)Chen, Wang, Chen, Wu, Liu, Chen, Li, Kanda, Yoshioka, Xiao, et~al.]{chen2022wavlm}
Chen, S., Wang, C., Chen, Z., Wu, Y., Liu, S., Chen, Z., Li, J., Kanda, N., Yoshioka, T., Xiao, X., et~al.
\newblock Wavlm: Large-scale self-supervised pre-training for full stack speech processing.
\newblock \emph{IEEE Journal of Selected Topics in Signal Processing}, 16\penalty0 (6):\penalty0 1505--1518, 2022.

\bibitem[Chen et~al.(2024)Chen, Ma, Yan, Liang, Li, Xu, Niu, Zhu, Yang, Liu, et~al.]{chen2024slam}
Chen, W., Ma, Z., Yan, R., Liang, Y., Li, X., Xu, R., Niu, Z., Zhu, Y., Yang, Y., Liu, Z., et~al.
\newblock Slam-omni: Timbre-controllable voice interaction system with single-stage training.
\newblock \emph{arXiv preprint arXiv:2412.15649}, 2024.

\bibitem[Chu et~al.(2023)Chu, Xu, Zhou, Yang, Zhang, Yan, Zhou, and Zhou]{chu2023qwen}
Chu, Y., Xu, J., Zhou, X., Yang, Q., Zhang, S., Yan, Z., Zhou, C., and Zhou, J.
\newblock Qwen-audio: Advancing universal audio understanding via unified large-scale audio-language models.
\newblock \emph{arXiv preprint arXiv:2311.07919}, 2023.

\bibitem[Chu et~al.(2024)Chu, Xu, Yang, Wei, Wei, Guo, Leng, Lv, He, Lin, et~al.]{chu2024qwen2}
Chu, Y., Xu, J., Yang, Q., Wei, H., Wei, X., Guo, Z., Leng, Y., Lv, Y., He, J., Lin, J., et~al.
\newblock Qwen2-audio technical report.
\newblock \emph{arXiv preprint arXiv:2407.10759}, 2024.

\bibitem[Copet et~al.(2024)Copet, Kreuk, Gat, Remez, Kant, Synnaeve, Adi, and D{\'e}fossez]{copet2024simple}
Copet, J., Kreuk, F., Gat, I., Remez, T., Kant, D., Synnaeve, G., Adi, Y., and D{\'e}fossez, A.
\newblock Simple and controllable music generation.
\newblock \emph{Advances in Neural Information Processing Systems}, 36, 2024.

\bibitem[Das et~al.(2024)Das, Dingliwal, Ronanki, Paturi, Huang, Mathur, Yuan, Bekal, Niu, Jayanthi, et~al.]{das2024speechverse}
Das, N., Dingliwal, S., Ronanki, S., Paturi, R., Huang, Z., Mathur, P., Yuan, J., Bekal, D., Niu, X., Jayanthi, S.~M., et~al.
\newblock Speechverse: A large-scale generalizable audio language model.
\newblock \emph{arXiv preprint arXiv:2405.08295}, 2024.

\bibitem[D{\'e}fossez et~al.(2022)D{\'e}fossez, Copet, Synnaeve, and Adi]{defossez2022high}
D{\'e}fossez, A., Copet, J., Synnaeve, G., and Adi, Y.
\newblock High fidelity neural audio compression.
\newblock \emph{arXiv preprint arXiv:2210.13438}, 2022.

\bibitem[D{\'e}fossez et~al.(2024)D{\'e}fossez, Mazar{\'e}, Orsini, Royer, P{\'e}rez, J{\'e}gou, Grave, and Zeghidour]{defossez2024moshi}
D{\'e}fossez, A., Mazar{\'e}, L., Orsini, M., Royer, A., P{\'e}rez, P., J{\'e}gou, H., Grave, E., and Zeghidour, N.
\newblock Moshi: a speech-text foundation model for real-time dialogue.
\newblock \emph{arXiv preprint arXiv:2410.00037}, 2024.

\bibitem[Du et~al.(2024)Du, Chen, Zhang, Hu, Lu, Yang, Hu, Zheng, Gu, Ma, et~al.]{du2024cosyvoice}
Du, Z., Chen, Q., Zhang, S., Hu, K., Lu, H., Yang, Y., Hu, H., Zheng, S., Gu, Y., Ma, Z., et~al.
\newblock Cosyvoice: A scalable multilingual zero-shot text-to-speech synthesizer based on supervised semantic tokens.
\newblock \emph{arXiv preprint arXiv:2407.05407}, 2024.

\bibitem[Fang et~al.(2024)Fang, Guo, Zhou, Ma, Zhang, and Feng]{fang2024llama}
Fang, Q., Guo, S., Zhou, Y., Ma, Z., Zhang, S., and Feng, Y.
\newblock Llama-omni: Seamless speech interaction with large language models.
\newblock \emph{arXiv preprint arXiv:2409.06666}, 2024.

\bibitem[Huang et~al.(2024)Huang, Li, Yang, Shi, Chang, Ye, Wu, Hong, Huang, Liu, et~al.]{huang2024audiogpt}
Huang, R., Li, M., Yang, D., Shi, J., Chang, X., Ye, Z., Wu, Y., Hong, Z., Huang, J., Liu, J., et~al.
\newblock Audiogpt: Understanding and generating speech, music, sound, and talking head.
\newblock In \emph{Proceedings of the AAAI Conference on Artificial Intelligence}, volume~38, pp.\  23802--23804, 2024.

\bibitem[Ji et~al.(2024)Ji, Chen, Fang, Zuo, Lu, Wang, Jiang, Zhou, Liu, Cheng, et~al.]{ji2024wavchat}
Ji, S., Chen, Y., Fang, M., Zuo, J., Lu, J., Wang, H., Jiang, Z., Zhou, L., Liu, S., Cheng, X., et~al.
\newblock Wavchat: A survey of spoken dialogue models.
\newblock \emph{arXiv preprint arXiv:2411.13577}, 2024.

\bibitem[Jiang et~al.(2023)Jiang, Sablayrolles, Mensch, Bamford, Chaplot, Casas, Bressand, Lengyel, Lample, Saulnier, et~al.]{jiang2023mistral}
Jiang, A.~Q., Sablayrolles, A., Mensch, A., Bamford, C., Chaplot, D.~S., Casas, D. d.~l., Bressand, F., Lengyel, G., Lample, G., Saulnier, L., et~al.
\newblock Mistral 7b.
\newblock \emph{arXiv preprint arXiv:2310.06825}, 2023.

\bibitem[Joshi et~al.(2017)Joshi, Choi, Weld, and Zettlemoyer]{joshi2017triviaqa}
Joshi, M., Choi, E., Weld, D.~S., and Zettlemoyer, L.
\newblock Triviaqa: A large scale distantly supervised challenge dataset for reading comprehension.
\newblock \emph{arXiv preprint arXiv:1705.03551}, 2017.

\bibitem[Kong et~al.(2020)Kong, Cao, Iqbal, Wang, Wang, and Plumbley]{kong2020panns}
Kong, Q., Cao, Y., Iqbal, T., Wang, Y., Wang, W., and Plumbley, M.~D.
\newblock Panns: Large-scale pretrained audio neural networks for audio pattern recognition.
\newblock \emph{IEEE/ACM Transactions on Audio, Speech, and Language Processing}, 28:\penalty0 2880--2894, 2020.

\bibitem[Lakhotia et~al.(2021)Lakhotia, Kharitonov, Hsu, Adi, Polyak, Bolte, Nguyen, Copet, Baevski, Mohamed, et~al.]{lakhotia2021generative}
Lakhotia, K., Kharitonov, E., Hsu, W.-N., Adi, Y., Polyak, A., Bolte, B., Nguyen, T.-A., Copet, J., Baevski, A., Mohamed, A., et~al.
\newblock On generative spoken language modeling from raw audio.
\newblock \emph{Transactions of the Association for Computational Linguistics}, 9:\penalty0 1336--1354, 2021.

\bibitem[Ma et~al.(2023)Ma, Zheng, Ye, Li, Gao, Zhang, and Chen]{ma2023emotion2vec}
Ma, Z., Zheng, Z., Ye, J., Li, J., Gao, Z., Zhang, S., and Chen, X.
\newblock emotion2vec: Self-supervised pre-training for speech emotion representation.
\newblock \emph{arXiv preprint arXiv:2312.15185}, 2023.

\bibitem[Ma et~al.(2024)Ma, Chen, Zhang, Zheng, Chen, Li, Ye, Chen, and Hain]{ma2024emobox}
Ma, Z., Chen, M., Zhang, H., Zheng, Z., Chen, W., Li, X., Ye, J., Chen, X., and Hain, T.
\newblock Emobox: Multilingual multi-corpus speech emotion recognition toolkit and benchmark.
\newblock \emph{arXiv preprint arXiv:2406.07162}, 2024.

\bibitem[Nachmani et~al.(2023)Nachmani, Levkovitch, Hirsch, Salazar, Asawaroengchai, Mariooryad, Rivlin, Skerry-Ryan, and Ramanovich]{nachmani2023spoken}
Nachmani, E., Levkovitch, A., Hirsch, R., Salazar, J., Asawaroengchai, C., Mariooryad, S., Rivlin, E., Skerry-Ryan, R., and Ramanovich, M.~T.
\newblock Spoken question answering and speech continuation using spectrogram-powered llm.
\newblock \emph{arXiv preprint arXiv:2305.15255}, 2023.

\bibitem[Nguyen et~al.(2025)Nguyen, Muller, Yu, Costa-Jussa, Elbayad, Popuri, Ropers, Duquenne, Algayres, Mavlyutov, et~al.]{nguyen2025spirit}
Nguyen, T.~A., Muller, B., Yu, B., Costa-Jussa, M.~R., Elbayad, M., Popuri, S., Ropers, C., Duquenne, P.-A., Algayres, R., Mavlyutov, R., et~al.
\newblock Spirit-lm: Interleaved spoken and written language model.
\newblock \emph{Transactions of the Association for Computational Linguistics}, 13:\penalty0 30--52, 2025.

\bibitem[Poria et~al.(2018)Poria, Hazarika, Majumder, Naik, Cambria, and Mihalcea]{poria2018meld}
Poria, S., Hazarika, D., Majumder, N., Naik, G., Cambria, E., and Mihalcea, R.
\newblock Meld: A multimodal multi-party dataset for emotion recognition in conversations.
\newblock \emph{arXiv preprint arXiv:1810.02508}, 2018.

\bibitem[Qin et~al.(2023)Qin, Cai, Jin, Yan, Liang, Zhu, Lin, Han, Ding, Wang, Xie, Qi, Liu, Sun, and Zhou]{qin2023webcpm}
Qin, Y., Cai, Z., Jin, D., Yan, L., Liang, S., Zhu, K., Lin, Y., Han, X., Ding, N., Wang, H., Xie, R., Qi, F., Liu, Z., Sun, M., and Zhou, J.
\newblock Webcpm: Interactive web search for chinese long-form question answering.
\newblock In \emph{Proceedings of ACL 2023}. Association for Computational Linguistics, 2023.
\newblock URL \url{https://arxiv.org/abs/2305.06849}.

\bibitem[Radford et~al.(2023)Radford, Kim, Xu, Brockman, McLeavey, and Sutskever]{radford2023robust}
Radford, A., Kim, J.~W., Xu, T., Brockman, G., McLeavey, C., and Sutskever, I.
\newblock Robust speech recognition via large-scale weak supervision.
\newblock In \emph{International conference on machine learning}, pp.\  28492--28518. PMLR, 2023.

\bibitem[Siuzdak et~al.(2024)Siuzdak, Gr{\"o}tschla, and Lanzend{\"o}rfer]{siuzdak2024snac}
Siuzdak, H., Gr{\"o}tschla, F., and Lanzend{\"o}rfer, L.~A.
\newblock Snac: Multi-scale neural audio codec.
\newblock In \emph{Audio Imagination: NeurIPS 2024 Workshop AI-Driven Speech, Music, and Sound Generation}, 2024.

\bibitem[Touvron et~al.(2023)Touvron, Lavril, Izacard, Martinet, Lachaux, Lacroix, Rozi{\`e}re, Goyal, Hambro, Azhar, et~al.]{touvron2023llama}
Touvron, H., Lavril, T., Izacard, G., Martinet, X., Lachaux, M.-A., Lacroix, T., Rozi{\`e}re, B., Goyal, N., Hambro, E., Azhar, F., et~al.
\newblock Llama: Open and efficient foundation language models.
\newblock \emph{arXiv preprint arXiv:2302.13971}, 2023.

\bibitem[Wang et~al.(2024{\natexlab{a}})Wang, Lu, Tang, Yan, Xia, and Xiong]{wang2024full}
Wang, P., Lu, S., Tang, Y., Yan, S., Xia, W., and Xiong, Y.
\newblock A full-duplex speech dialogue scheme based on large language models.
\newblock \emph{arXiv preprint arXiv:2405.19487}, 2024{\natexlab{a}}.

\bibitem[Wang et~al.(2024{\natexlab{b}})Wang, Li, Fu, Xie, Li, Sun, and Ma]{wang2024freeze}
Wang, X., Li, Y., Fu, C., Xie, L., Li, K., Sun, X., and Ma, L.
\newblock Freeze-omni: A smart and low latency speech-to-speech dialogue model with frozen llm.
\newblock \emph{arXiv preprint arXiv:2411.00774}, 2024{\natexlab{b}}.

\bibitem[Xie \& Wu(2024)Xie and Wu]{xie2024mini}
Xie, Z. and Wu, C.
\newblock Mini-omni: Language models can hear, talk while thinking in streaming.
\newblock \emph{arXiv preprint arXiv:2408.16725}, 2024.

\bibitem[Yang et~al.(2024)Yang, Yang, Zhang, Hui, Zheng, Yu, Li, Liu, Huang, Wei, et~al.]{yang2024qwen2}
Yang, A., Yang, B., Zhang, B., Hui, B., Zheng, B., Yu, B., Li, C., Liu, D., Huang, F., Wei, H., et~al.
\newblock Qwen2. 5 technical report.
\newblock \emph{arXiv preprint arXiv:2412.15115}, 2024.

\bibitem[Zeng et~al.(2024{\natexlab{a}})Zeng, Du, Liu, Wang, Jiang, Zhao, Dong, and Tang]{zeng2024glm}
Zeng, A., Du, Z., Liu, M., Wang, K., Jiang, S., Zhao, L., Dong, Y., and Tang, J.
\newblock Glm-4-voice: Towards intelligent and human-like end-to-end spoken chatbot.
\newblock \emph{arXiv preprint arXiv:2412.02612}, 2024{\natexlab{a}}.

\bibitem[Zeng et~al.(2024{\natexlab{b}})Zeng, Du, Liu, Zhang, Jiang, Dong, and Tang]{zeng2024scaling}
Zeng, A., Du, Z., Liu, M., Zhang, L., Jiang, S., Dong, Y., and Tang, J.
\newblock Scaling speech-text pre-training with synthetic interleaved data.
\newblock \emph{arXiv preprint arXiv:2411.17607}, 2024{\natexlab{b}}.

\bibitem[Zhao et~al.(2022)Zhao, Zhang, Hu, Liu, Jin, Wang, and Li]{zhao2022m3ed}
Zhao, J., Zhang, T., Hu, J., Liu, Y., Jin, Q., Wang, X., and Li, H.
\newblock M3ed: Multi-modal multi-scene multi-label emotional dialogue database.
\newblock \emph{arXiv preprint arXiv:2205.10237}, 2022.

\end{thebibliography}
\bibliographystyle{icml2025}

\newpage


\end{document}